\documentclass{article}

\usepackage{microtype}
\usepackage{graphicx}
\usepackage{subfigure}
\usepackage{booktabs}
\usepackage{hyperref}
\usepackage{caption}
\usepackage{multirow}

\usepackage[accepted]{icml2023}

\usepackage{amsmath}
\usepackage{amssymb}
\usepackage{mathtools}
\usepackage{amsthm}
\usepackage{amsmath}
\usepackage{algorithm}
\usepackage{algorithmicx}
\usepackage[noend]{algpseudocode}

\usepackage[capitalize,noabbrev]{cleveref}

\theoremstyle{plain}

\theoremstyle{definition}

\theoremstyle{remark}

\usepackage[textsize=tiny]{todonotes}

\usepackage{xcolor}
\usepackage{array}
\usepackage[export]{adjustbox}
\usepackage{balance}

\newcommand{\papertitle}{ObjectLab: Automated Diagnosis of Mislabeled Images in Object Detection Data}

\icmltitlerunning{\papertitle}

\begin{document}

\twocolumn[
\icmltitle{\papertitle}



\icmlsetsymbol{equal}{*}

\begin{icmlauthorlist}
\icmlauthor{Ulyana Tkachenko}{equal,comp}
\icmlauthor{Aditya Thyagarajan}{equal,comp}
\icmlauthor{Jonas Mueller}{comp}
\end{icmlauthorlist}
\icmlaffiliation{comp}{Cleanlab}
\icmlcorrespondingauthor{Jonas Mueller}{jonas@cleanlab.ai}

\icmlkeywords{Machine Learning, ICML}

\vskip 0.3in
]



\printAffiliationsAndNotice{\icmlEqualContribution} 

\begin{abstract}
Despite powering sensitive systems like autonomous vehicles, object detection remains fairly brittle in part due to annotation errors that plague most real-world training datasets. 
We propose ObjectLab, a straightforward algorithm to detect diverse errors in object detection labels, including: overlooked bounding boxes, badly located boxes, and incorrect class label assignments. ObjectLab utilizes \emph{any} trained object detection model to score the label quality of each image, such that mislabeled images can be automatically prioritized for label review/correction. Properly handling erroneous data enables training a better version of the same object detection model, without any change in existing modeling code. Across different object detection datasets (including COCO) and different models (including Detectron-X101 and Faster-RCNN), ObjectLab consistently detects annotation errors with much better precision/recall compared to other label quality scores.
\end{abstract}

\section{Introduction}
\label{introduction}
Object Detection is a key computer vision task powering many high-impact applications where computers decide actions based on captured images via a learned model. The datasets used to train/evaluate these detectors require a massive amount of human labeling which is inevitably imperfect. Annotators of an object detection dataset inspect an image and, for each depicted object, are supposed to draw a bounding box around it and assign a discrete class label to categorize this object.

\begin{figure}[h!]
\begin{center}
\includegraphics[width=\columnwidth+0.4cm]{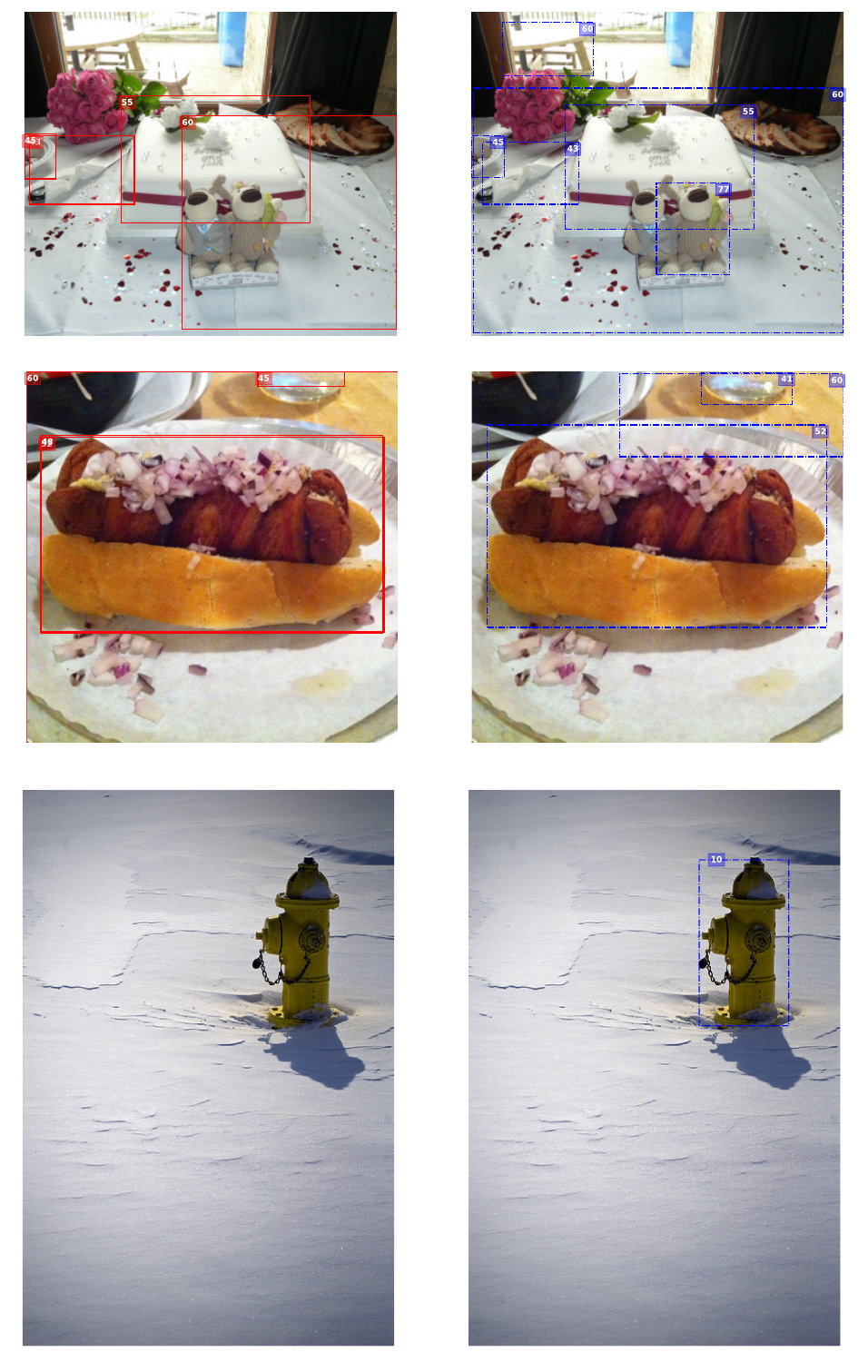} \\
\vspace*{-1.5em}
\caption{Images in COCO dataset with some of the \emph{lowest} ObjectLab label quality scores. We show both the original given label (left column in \textcolor{red}{red}) and prediction from Detectron2-X101 model (right column in \textcolor{blue}{blue}). 
These examples exhibit different naturally-occurring label errors: (top) \emph{Badly Located} box, (middle) \emph{Swapped} box, (bottom) \emph{Overlooked} class label.
In top row: annotators poorly outlined only half of the \textbf{dinning table} (class \#60) which the model localized much better (with confidence 0.964), leading to a low Badly-Located score in ObjectLab.
In middle row: the glass object on the right is incorrectly annotated as a \textbf{bowl} (class \#45), while the model predicted \textbf{cup} (class \#41) with confidence 0.962, leading to a low Swapped-score in ObjectLab.
In bottom row: annotators missed the \textbf{fire hydrant} (class \#10 in COCO) which the model detected with confidence 0.998, leading to a low Overlooked-score in ObjectLab.
}
\label{fig:errortypes}
\end{center}
\end{figure}

In real-world datasets, annotators make three types of mistakes depicted in Figure \ref{fig:errortypes}: (1) An \textbf{Overlooked} error in which a depicted object was not spotted and thus no corresponding bounding box around it exists in the given label for this image, (2) A \textbf{Badly Located} error in which annotators sloppily draw the bounding box around a depicted object such that its location/edges fail to perfectly enclose the object, (3) A \textbf{Swapped} error in which annotators draw a correct bounding box around a depicted object, but assign it the wrong class label. Such \emph{Swapped} errors are also common in many classification datasets \cite{northcutt2021pervasive}, but the increased complexity of object detection annotation introduces potential for more varied types of label errors than encountered in classification. 
We propose an algorithm, \textbf{ObjectLab}, that utilizes \emph{any} trained \textbf{object} detection model to estimate the incorrect \textbf{lab}els in such a dataset, regardless which of these 3 types of mistake the data annotators made.

Training and evaluating models with incorrect bounding box annotations is clearly worrisome. Likely mislabeled images in an object detection dataset should be reviewed and either re-labeled or excluded from the dataset. While some research advocates dealing with noisy labels by training models in a special manner \cite{nishi2021augmentation,sukhbaatar2014learning,jiang2018mentornet,zhang2018generalized}, we advocate for a straightforward data-centric approach to improve the data directly by first estimating which images are mislabeled. The data-centric approach has many advantages over special modeling \cite{kuan,northcutt2021confident} -- most importantly, it can be used to improve the performance of \emph{any} object detection model, regardless of its architecture or training strategy. Our ObjectLab approach\footnote{Code to run   our method: \\ \url{https://github.com/cleanlab/cleanlab} \\  Code to reproduce benchmarks: \url{https://github.com/cleanlab/object-detection-benchmarks}} utilizes \emph{any} trained detector to estimate \emph{which} images are incorrectly labeled -- these can be corrected to subsequently produce an even better version of this same model (without any change in the existing modeling code). This generality ensures data-centric methods like ObjectLab will remain valuable in the future computer vision toolkit, even when new architectures and training strategies have been invented which invalidate special techniques developed specifically for today's models.

\section{Methods}
\label{methods}

Our aim is detecting labeling issues in a standard object detection dataset, in which each image $I$ is annotated with bounding boxes $B$ around each depicted object, and each box is given an corresponding class label $c(B) \in \{1,\dots,K\}$ categorizing the object into one of $K$ classes. We interchangeably refer to the set of bounding boxes provided for image $I$ in the original dataset, $\mathcal{L}(I)$, as the \emph{given label} or \emph{annotation} for this image.

Here we estimate a \textbf{label quality score} $\hat{s}(I)$ for each image, such that images receiving lower scores are more likely mislabeled \cite{kuan}, suffering from at least one of the aforementioned mistake-types (1)-(3). Practical constraints limit the number of images whose label can be reviewed, and thus an effective label quality score is key to prioritizing which images are worth another look. Good scoring methods ensure reviewers neither waste time inspecting correctly labeled images (high \emph{precision}) nor fail to catch images with label errors (high \emph{recall}).

A natural way to score label quality is via the predictions of a ML model. In this paper, we restrict ourselves to scoring methods that can be applied to the predictions from \emph{any} standard object detection model, no matter its architecture or training strategy. A key question is:  \emph{when to trust the model over the original data}?
To avoid bias from overfitting, label quality scores are computed using \emph{out-of-sample} predictions from the model, i.e.\ based on predictions for an image $I$ from a copy of the model which never saw $I$ during training. We obtain out-of-sample predictions for every image in a dataset via straightforward $5$-fold cross-validation.

Given an image $I$, a typical object detection model outputs \emph{prediction} $\hat{\mathcal{P}}(I)$, which is a set of \emph{predicted bounding boxes} $\hat{B}$, each localizing an estimated object and associated with: \emph{predicted class} $\hat{c}(\hat{B}) \in \{1, \dots, K\}$ and a \emph{confidence} value $0 \le \hat{p}(\hat{B}) \le 1$ reflecting the estimated probability that the object localized by $\hat{B}$ belongs to class  $\hat{c}(\hat{B})$. 
The set $\hat{\mathcal{P}}(I)$ is typically only comprised of predicted boxes $\hat{B}$ whose confidence exceeds some fixed threshold, i.e.\ $\hat{p}(\hat{B}) > \tau_{\downarrow}$. Here we use the same value $\tau_{\downarrow} = 0.5$ as the default in many popular object detection libraries, noting the empirical performance of ObjectLab was not significantly affected by trying smaller values of $\tau_{\downarrow}$ in our benchmarks. 
While certain types of object detection models output richer information than listed here \cite{zaidi2022survey}, we believe being compatible with \emph{any} type of model is key to an effective label quality score for data-centric AI. As object detection architectures advance over time and models become more accurate/calibrated, such label quality scores will remain applicable and immediately detect errors more effectively.

\subsection{ObjectLab}
\label{methods:objectlab}

Our proposed \textbf{ObjectLab} method straightforwardly scores each image $I$ independently of the others and produces a label quality score $\hat{s}(I)$ solely based on the given label $\mathcal{L}(I)$ and model prediction $\hat{\mathcal{P}}(I)$. Algorithm \ref{alg:objectlab} details our method. 
The ObjectLab score is a holistic representation of all possible labeling errors that can occur and is based on a geometric mean of three mistake-subtype scores $\hat{s}_{overlook}(I), \hat{s}_{badloc}(I), \hat{s}_{swap}(I)$ which respectively evaluate how likely this image suffers from an \emph{Overlooked}, \emph{Badly Located}, or \emph{Swapped} error. These mistake-subtype scores are each computed by estimating a particular aspect of each annotated/predicted bounding box in image $I$ via a particular quality score in [0,1], and subsequently pooling these quality estimates over all of the relevant boxes in $I$ to form one of the subtype scores for $I$.

To gain some intuition, consider say the \emph{Badly located} mistake-subtype, for which we compute a particular quality estimate for each annotated bounding box $B$ in $I$, reflecting the quality of its location.  These location-quality estimates are then pooled over every annotated box $B$ in $I$ to form the single subtype score $\hat{s}_{badloc}(I)$, which roughly quantifies the estimated likelihood that any box $B$ in $I$ was badly located. Pooling via the mean quality estimate is overly sensitive to statistical variation the quality estimates for correctly located boxes, while pooling via the minimum quality estimate undesirably ignores the estimates for all boxes except one. An effective compromise between these extremes is \emph{softmin} pooling \cite{eric}, in which we compute the pooled value $0 \le \bar{q} \le 1$ from a vector of per-box values $\vec{q}: = \langle q_1, ..., q_N \rangle$ via the inner product: $\bar{q} = \langle \vec{q} , \text{softmax}(1-\vec{q}) \rangle$. Here we suppose there are $N$ annotated boxes for $I$. Such pooling reflects a softer version of the minimum function that still takes all boxes' quality estimates into account. Beyond the \emph{Badly located} mistake-subtype, we also employ softmin pooling to aggregate certain per-box quality estimates into the other two types of subtype scores $\hat{s}_{overlook}(I), \hat{s}_{swap}(I)$.

\begin{algorithm}[tb]
\caption{ObjectLab score $\hat{s}(I)$ for an image $I$}
\label{alg:objectlab}
\begin{algorithmic}[1]
\Require{given label $\mathcal{L}(I)$, model prediction $\hat{\mathcal{P}}(I)$}
 \State \raggedright $q_1,\dots,q_N \ \gets$ \ BadlocBoxScores($\mathcal{L}(I), \hat{\mathcal{P}}(I)$)
 \State $\hat{s}_{badloc}(I) \ \gets$ \ \text{softmin}($q_1,\dots,q_N$)
  \State \raggedright $q_1,\dots,q_N \ \gets$ \ SwappedBoxScores($\mathcal{L}(I), \hat{\mathcal{P}}(I)$)
 \State $\hat{s}_{swap}(I) \ \gets$ \ \text{softmin}($q_1,\dots,q_N$)
  \State \raggedright $q_1,\dots,q_M \ \gets$ \ OverlookedBoxScores($\mathcal{L}(I), \hat{\mathcal{P}}(I)$)
 \State $\hat{s}_{overlook}(I) \ \gets$ \ \text{softmin}($q_1,\dots,q_M$) \\
\hspace*{-6mm} \Return Geometric mean of $\hat{s}_{badloc}, \hat{s}_{swap}, \hat{s}_{overlook}$ for $I$
\end{algorithmic}
\end{algorithm}

Algorithms \ref{alg:badloc}, \ref{alg:swapped}, \ref{alg:overlooked} detail the computation of the per-box quality estimates used for each subtype score. 
Intuitively, our quality score for an individual annotated box being badly located is based on its similarity with the nearest predicted box of the same class. Our quality score for an individual annotated box having a swapped class label is inversely related to its similarity with a nearby predicted box confidently predicted to belong to a different class. To estimate whether an individual predicted box $\hat{B}$ corresponds to an overlooked box that should have been in the original annotations, our corresponding quality score is based on  this prediction's confidence and the similarity between $\hat{B}$ and the nearest annotated box of the same class as $\hat{c}(\hat{B})$.

Between any pair of bounding boxes in the same image, we define a \textbf{similarity} function: 
$$
sim(B_1, B_2) = \alpha \cdot k(B_1, B_2) + (1-\alpha) \cdot IoU(B_1, B_2).
$$ 
Here IoU is the standard \emph{Intersection over Union} similarity measure and $k(\cdot, \cdot)$ is a Gaussian kernel similarity between 4D vectors defined by the outer edges of each box = $\exp ( - ||\vec{b}_1 - \vec{b}_2|| / \sigma)$ where the entries of $\vec{b}_1$ (or $\vec{b}_2$) are the coordinates of the top-left and bottom-right corners of $B_1$ (or $B_2$) normalized to unit interval. Throughout we simply set $\alpha = \sigma = 0.1$ and did not find their precise values to affect performance. Rather the Gaussian kernel is included to avoid similarity ties when the IoU is equal to 0. Tied label quality scores between different images are undesirable as they do not aid in prioritizing which to review first. We also define $sim_*$ as the \textbf{minimum possible similarity} between any pair of annotated and predicted boxes across all images in the dataset, and $q^* = 1$ as the \textbf{maximum possible quality estimate} for any box (across all mistake subtypes).

\paragraph{Computing Badly Located scores per annotated box.}
\label{methods:badloc}

Badly located box scores are calculated for every annotated box via Algorithm \ref{alg:badloc}. 
We simply score each annotated box based on its similarity with the nearest predicted box of the same class that shares some overlap. When there is no such predicted box, we consider this annotated box well-located (assigning maximum quality score $q^* = 1$). For a well-trained model, we observed the majority of incorrect  predictions are entirely false positive/negative detections; when objects are correctly detected, their predicted bounding boxes tend to be well-localized, except for classes where the original annotations also contain poorly located boxes thus confusing the model.

\paragraph{Computing Swapped scores per annotated box.}
\label{methods:swap}

Swapped box scores are calculated for every annotated box via Algorithm \ref{alg:swapped}.
We are most concerned that an annotated box may have a swapped class label when there exists an extremely similar predicted box that was predicted with high confidence to belong to a different class. For swapped errors, we score the annotated box quality based on the distance to the most similar predicted box that was confidently predicted to belong to a different class. If there are no such predicted boxes, we do not consider the annotated box to potentially have a swapped class label, and its quality estimate is set to the maximum value $q^* = 1$.  
Deciding what constitutes a highly confident prediction depends on a fixed threshold $\tau_{\uparrow}$, which one can set based on the estimated trustworthiness of the model (e.g.\ via a calibration curve). One can also use separate thresholds for each class \cite{northcutt2021confident}. Here we simply fix this threshold $\tau_{\uparrow} = 0.95$ corresponding to a 95\% confidence value adopted as a de facto standard in statistical decision-making. 

\paragraph{Computing Overlooked scores per predicted box.}
\label{methods:overlooked}
While the aforementioned quality estimates for the \emph{Swapped} and \emph{Badly Located}  error types are computed for each annotated box, \emph{Overlooked} errors are defined by the absence of such a box in the given label.
Thus overlooked box quality scores are instead calculated for every predicted box via Algorithm \ref{alg:overlooked}.  Again we only consider high-confidence predicted boxes whose confidence exceeds threshold $\tau_{\uparrow}$. For each such box, we consider whether there is a corresponding annotated box with no overlap present in the image. If there is, the similarity between the two serves as the quality score. Otherwise we use the minimum possible similarity adjusted by the model confidence (since a 99\% confident prediction with no corresponding annotated box is more indicative of an overlooked error than a 98\% confident prediction).

\begin{algorithm}[tb]
\caption{BadlocBoxScores for image $I$}\label{alg:badloc}
\begin{algorithmic}[1]
\Require{given label $\mathcal{L}(I)$, model prediction $\hat{\mathcal{P}}(I)$}
 \State \raggedright \texttt{scores} $\gets \{\}$
 \For{annotated box $B \in \mathcal{L}(I)$}
  \State Let $k = c(B)$ denote its annotated class and \\
  \hskip\algorithmicindent $\mathcal{P}_k := \{ \hat{B} \in \hat{P}(I): \hat{c}(\hat{B}) = k, IoU(\hat{B}, B) > 0\}$ \par 
  \hskip\algorithmicindent denote the predicted boxes with the same \par
  \hskip\algorithmicindent predicted class that overlap with $B$.
  \State \texttt{scores.append}($q$) \ \ where $ q \gets q^*$ \textbf{if} $\mathcal{P}_k = \emptyset$ \par
  \hskip\algorithmicindent \hspace*{23mm} \textbf{else}: $q \gets \max_{\hat{B} \in \mathcal{P}_k} sim(B, \hat{B})$
 \EndFor
 \Return \texttt{scores}
\end{algorithmic}
\end{algorithm}

\begin{algorithm}[tb]
\caption{SwappedBoxScores for image $I$}\label{alg:swapped}
\begin{algorithmic}[1]
\Require{given label $\mathcal{L}(I)$, model prediction $\hat{\mathcal{P}}(I)$}
 \State \raggedright \texttt{scores} $\gets \{\}$
 \For{annotated box $B \in \mathcal{L}(I)$}
   \State Let $k = c(B)$ denote its annotated class and  \par
  \hskip\algorithmicindent $\mathcal{P}_{-k} := \{ \hat{B} \in \hat{P}(I): \ \hat{c}(\hat{B}) \neq k, \ \hat{p}(\hat{B}) > \tau_{\uparrow} \}$ be \par
  \hskip\algorithmicindent  the predicted boxes with another predicted class  \par
  \hskip\algorithmicindent  whose confidence exceeds threshold $\tau_{\uparrow}$.
  \State \texttt{scores.append}($q$) \ \ where $ q \gets q^*$ \textbf{if} $\mathcal{P}_{-k} = \emptyset$ \par
  \hskip\algorithmicindent \hspace*{12mm} \textbf{else}: $q \gets  1 - \max_{\hat{B} \in \mathcal{P}_{-k}} sim(B, \hat{B}) $
 \EndFor
 \Return \texttt{scores}
\end{algorithmic}
\end{algorithm}

\begin{algorithm}[t]
\caption{OverlookedBoxScores for image $I$}\label{alg:overlooked}
\begin{algorithmic}[1]
\Require{given label $\mathcal{L}(I)$, model prediction $\hat{\mathcal{P}}(I)$}
 \State \raggedright \texttt{scores} $\gets \{\}$;
 \For{predicted box $\hat{B} \in  \hat{\mathcal{P}}(I)$ with $\hat{p}(\hat{B}) > \tau_{\uparrow}$}
  \State Let $k = \hat{c}(\hat{B})$ denote its predicted class and 
   \par \hskip\algorithmicindent 
  $\mathcal{L}_k := \{ B \in \mathcal{L}(I) : c(B) = k,  IoU(B, \hat{B}) = 0\}$ \par
   \hskip\algorithmicindent denote the subset of annotated boxes for\par 
   \hskip\algorithmicindent class $k$ that do not overlap with $\hat{B}$.
  \If{$\mathcal{L}_k = \emptyset$}
   \State $q = sim_* \cdot \big( 1 - \hat{p}(\hat{B})\big)$
  \Else
   \State $q = \max_{B \in \mathcal{L}_k} sim(B, \hat{B})$
  \EndIf
  \State \texttt{scores.append}($q$) 
 \EndFor
 \State \textbf{if} \texttt{scores} $=\emptyset$: \ \ \texttt{scores} $= \{ q^* \}$ \\
 \Return \texttt{scores}
\end{algorithmic}
\end{algorithm}

Aggregating separately quantified evidence for these various types of potential errors, the 
ObjectLab score offers a model-agnostic and computationally-efficient estimate of label quality in object detection datasets. This method straightforwardly utilizes predictions from a trained model. Sorting images by our proposed score can help detect a wide variety of labeling errors of different types. 

\section{Related Work}
\label{baseline_method}

Several previous works have demonstrated object detection datasets are full of labeling errors, mostly via manual analysis \cite{murrugarra2022can, baobab, sama, hasty}.
Other work has focused only on specific error types in object detection data and model-specific  techniques to improve training with noisily labeled data \cite{xu2019missing}.
Due to the practical value of systematic label error detection, extensive research on this task has been conducted particularly for classification datasets \cite{brodley1999identifying, Muller_2019}. \emph{Confident Learning} \cite{northcutt2021confident} is one particularly popular methodology to automatically detect mislabeled classification data. Recent work has studied methods to extend these label error detection capabilities beyond classification to structured data in NLP  \cite{klie2022annotation} and segmentation data in computer vision  \cite{rottmann2022automated, sergmentmeifyoucan}. In these areas, label quality scores have been found to be effective \cite{kuan}, particularly when they are properly pooled for data with complex multi-dimensional labels, e.g.\ via the softmin pooling used in ObjectLab \cite{eric, thyagarajan2022identifying}.

Some of the related label quality scoring methods discussed in this section are considered as baseline methods for comparison in our subsequent benchmarks. We focus on general approaches that can be used with any standard model and training strategy (like ObjectLab), because the score is solely based on model predictions and the original labels.  Throughout all such methods are only applied to out-of-sample predictions obtained via cross-validation.

\subsection{mAP label quality score}
\label{baseline_method:mAP}

In \emph{error analysis}, one sorts the data by the predictions' loss according to a standard evaluation metric computed separately for each instance. The resulting ranking reveals images for which the model struggles most, often because some of them are mislabeled \cite{bolya2020tide, Voxel51, klie2022annotation}. 
Thus this constitutes a reasonable approach to label quality scoring for general ML tasks.

\emph{Mean Average Precision} (mAP) is a widely used evaluation metric in object detection, quantifying the accuracy of an object detector via both its precision (percentage of correctly identified objects out of all the predicted objects) and recall (percentage of correctly detected objects out of all the annotated objects). mAP is computed by calculating the Average Precision (AP) for each class and then taking the mean across all classes.
As a standard approach to error analysis in object detection, we can compute mAP per image, and also use this as an alternative label quality score \cite{Voxel51}. 
If one had a perfect model, label quality scoring via mAP would properly measure how wrong data annotations are. However model predictions are never  perfect, especially with noisily labeled training datasets.

\subsection{Tile-estimates approach}
\label{baseline_method:Tile-estimates}

Beyond error analysis with per-image mAP scores, we also consider a direct extension of  methods for identifying label errors in classification tasks \cite{northcutt2021confident,kuan} to the object detection setting. 
To do so, we simply reduce aspects of object detection to a classification perspective. A straightforward way is a tile-based reduction, in which we first divide each image into a grid of size $(J, J)$. Each tile in this grid is assigned a label and predicted class probabilities, and across all the images, these tiles are treated as separate instances in a classification task (ignoring which image each tile stems from). Subsequently standard label-quality scoring for classification \cite{kuan} can be applied to score each tile. Here we use the likelihood of the given label according to the predicted class probabilities. We pool the tile-scores over an image to get its label quality score. We explored various pooling options and a simple geometric mean worked well.

To assign labels for each tile based on the given object detection annotations, we compute the overlap between tiles and annotated bounding boxes and assign the original bounding box label to the tiles significantly overlapping with this box. To obtain predicted class probabilities for a tile from our object detection model outputs, we form a kernel-smoothing predictor within each image. Here we apply the aforementioned similarity function $sim()$ between boxes but here to a tile and each predicted bounding box, in order to construct a similarity-weighted average of all boxes probabilities (for this same image) as the predicted class probability vector for the tile.

\subsection{CLOD \cite{chachula2022combating}}
\label{baseline_method:CLOD}

Similar to our above extension of classification label quality scores to the object setting, \citet{chachula2022combating} propose an extension of the \emph{Confident Learning} \cite{northcutt2021confident} approach to detect label errors in object detection. Their CLOD method involves clustering annotated and model-predicted boxes based on IoU distance -- specifically single linkage agglomerative clustering. As in our tile-estimates approach, CLOD assigns a label and predicted probabilities to each cluster, which allows the application of Confident Learning to assess label quality (treating each cluster as a separate instance).
Subsequently, one can simply use mean-pooling over the clusters within an image to obtain a label quality score for the image. 
\citet{chachula2022combating} also consider the Overlooked, Swapped, and Badly Located object detection errors and propose CLOD as an effective way to detect them. 


\section{Experiments}
\label{experiments}

\subsection{Dataset and Models}
\label{dataset_models}
Our benchmarks evaluate label quality scoring methods by training two object detection models across three datasets in order to ensure the results are dataset and model agnostic. Table \ref{tab:accuracies} reports the accuracy of each model on each dataset. 
Our first two datasets (COCO-bench and SYNTHIA) are specially curated for evaluation by ensuring we know which images are truly mislabeled or not.

\begin{figure}[tb]
\vskip 0.2in
\begin{center}
\centerline{\includegraphics[width=\columnwidth]{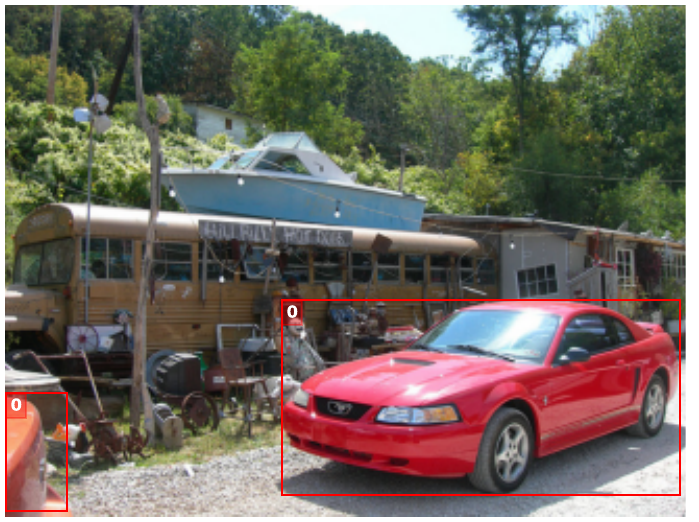}}
\caption{Example of a naturally mislabeled image in the COCO-bench dataset that receives low ObjectLab score. We show the original dataset label, which contains no boxes beyond those depicted for \textbf{Car} (class \#0).
Here the seated \textbf{person} and the \textbf{chair} were overlooked in the given label, even though these are among the 5 COCO-bench classes.}
\label{fig:cocobench}
\end{center}
\vskip -0.2in
\end{figure}

\paragraph{COCO-bench dataset.} This subset of 2171 images from the famous COCO 2017 dataset \cite{coco} only considers 5 of the classes: person, chair, cup, car, and traffic light.
These images and classes were selected because two other groups have re-annotated these images considering these classes \cite{baobab, sama}, and we use these redundant annotations here to determine which of these images truly have an annotation error in COCO. Specifically for each image, we compared its 5-class COCO annotation to its independent annotation by \citet{baobab} and by \citet{sama} in order to determine ground truth. When both of these extra annotations disagreed with the COCO annotation but agreed with one another, we considered the image to be mislabeled in our COCO-bench dataset (whose labels all come from 5-class COCO, not the extra annotations). When all three annotations agreed, we considered the image correctly labeled. Here \emph{agreement} between annotations was assessed by thresholding their pairwise mAP score. 
We manually inspected the remaining images to decide which were mislabeled or not. In total, 251 images in COCO-bench are considered mislabeled and we are confident in the ground truth assessments. Figure \ref{fig:cocobench} depicts an example from this benchmark that this process revealed to be truly mislabeled.

\begin{figure}[tb]
\begin{center}
\includegraphics[width=\columnwidth/2]{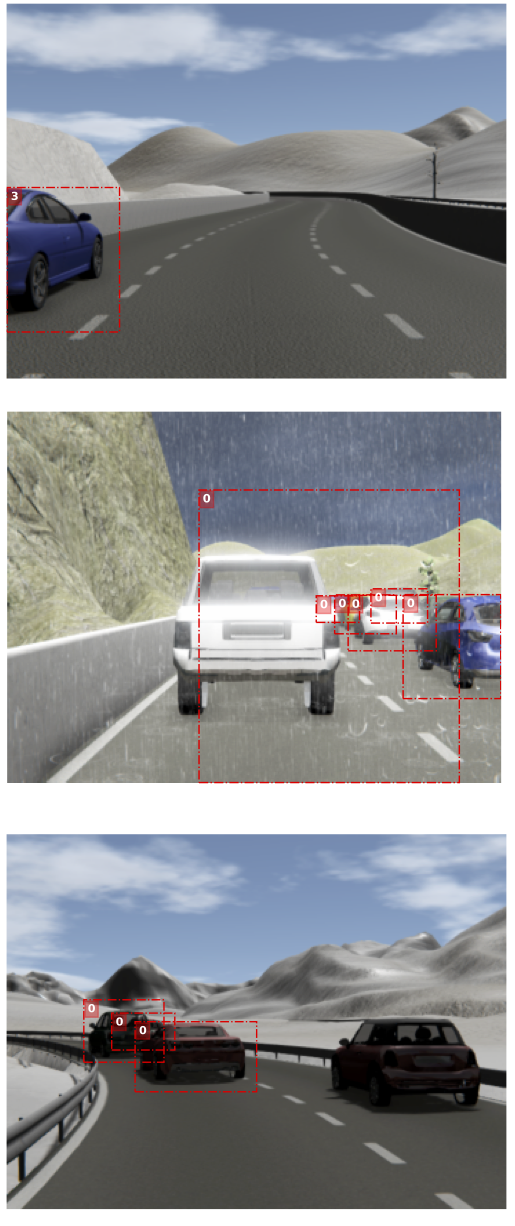}
\caption{Various errors in our SYNTHIA-AL dataset, including: class label for \textbf{Car} (class \#0 in SYNTHIA-AL) swapped with \textbf{Bicycle} (class \#3), bounding box around depicted  \textbf{Car} shifted to incorrect location (middle), and omitted bounding box around depicted \textbf{Car} (bottom). These examples involve the \textbf{Car} class, but similar errors exist in labels for each of the other 4 classes.}
\label{fig:synthia}
\end{center}
\end{figure}

\paragraph{SYNTHIA-AL dataset.} \citet{synthial} curated the SYNTHIA-AL dataset as an object detection benchmark where the ground truth labels are known because the images are synthetically generated by a realistic graphics engine \cite{German}. To ensure more independent images from what was originally a video dataset, we ensured a minimum distance of 28 frames between any pair of images included in our benchmark dataset. Our benchmark version of this  dataset contains 5000 images and 5 classes: Pedestrian, TrafficLight, Car, TrafficSign, Bicycle. 

We then randomly perturbed some of the clean labels in this dataset to inject various types of mislabeling: dropped bounding boxes, swapped class labels, and shifted bounding boxes. Some images contained more than one type of annotation error with 22\% unique images containing at least one error. Our perturbations were considered as sole source of ground truth error for the SYNTHIA-AL dataset, given the images are generated by a graphics engine. Figure \ref{fig:synthia} depicts some examples from this benchmark. 
While we cannot characterize all properties of the naturally-occurring label errors present in COCO-bench, we control the label errors in SYNTHIA-AL, facilitating more systematic evaluation.  

\paragraph{COCO-full dataset.} 

Finally we also considered detecting mislabeled images in the full COCO 2017 training dataset, which has 80 classes \cite{coco}. We refer to this dataset of 118,000 images as \emph{COCO-full} to distinguish it from COCO-bench. Figure \ref{fig:errortypes} shows some of the label errors automatically detected in this dataset. Note that we do know the ground truth mislabeled images in COCO, and thus only performed a limited evaluation with this dataset. For each label quality scoring method, we manually reviewed its 100 lowest-scoring images in COCO to assess what fraction of them were actually mislabeled.

\paragraph{Detectron-X101 model.} The Detectron-X101 network is one of the most accurate object detection models in the popular Detectron2 library \cite{wu2019detectron2}. This model uses a ResNeXt backbone \cite{resnext} with a  Feature Pyramid Network \cite{fpn}; standard convolutional and fully-connected output heads are used for box prediction.

\paragraph{Faster-RCNN model.} Proposed by \citet{ren2015faster}, the Faster R-CNN architecture is one of the most widely used object detection methods. Here we specifically use the R-50-FPN Faster-RCNN network from the MMDetection library \cite{mmdetection}.
This model shares parameters between a fully-convolutional region proposal network and the detection network, which is based on a ResNet-50 \cite{resnet} backbone with a Feature Pyramid Network \cite{fpn}. Faster-RCNN is slightly less accurate than Detectron-X101 (Table \ref{tab:accuracies}) but often favored for its efficiency. \looseness=-1

\subsection{Evaluation Metrics}
\label{eval_metrics}

Our primary interest is how well our label quality estimates correctly prioritize images that have annotation errors over those which do not. Label error detection can be viewed as a form of information retrieval, a field with standard evaluation metrics based around precision/recall, which we adopt here to evaluate different label quality scoring methods \cite{kuan}. For each set of label quality scores, we compare them against the ground truth information about which images are mislabeled to 
compute their \emph{Average Precision}, \emph{Precision @ 100} (i.e.\ what fraction of the 100 lowest-scoring images are truly mislabeled), and \emph{Precision @} $T$, for $T = $ the true number of mislabeled images in each dataset. For the full COCO dataset where this ground truth information is not available, we only report \emph{Precision @ 100} for a select number of label quality scoring methods (as it is labor intensive to report).


\section{Results}
\label{results}

\begin{figure*}[tb]
\centering
\begin{tabular}{cc}
  \includegraphics[width=85mm,valign=c]{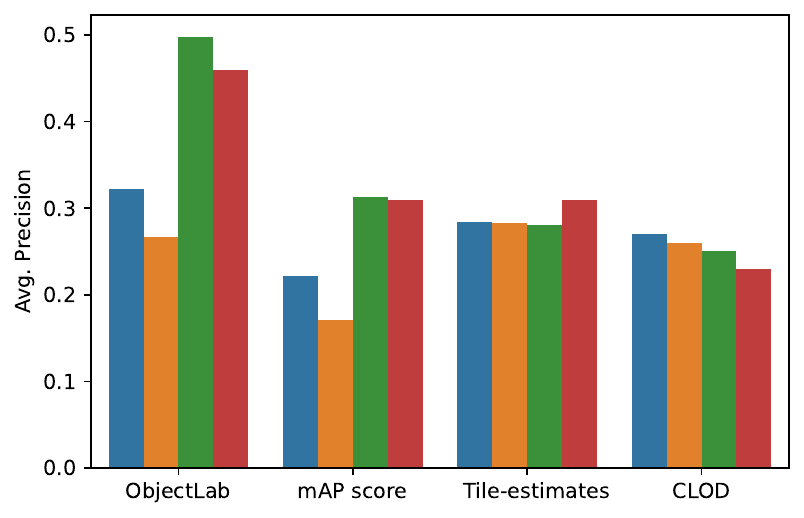} & \hfill  \includegraphics[width=70mm,valign=c]{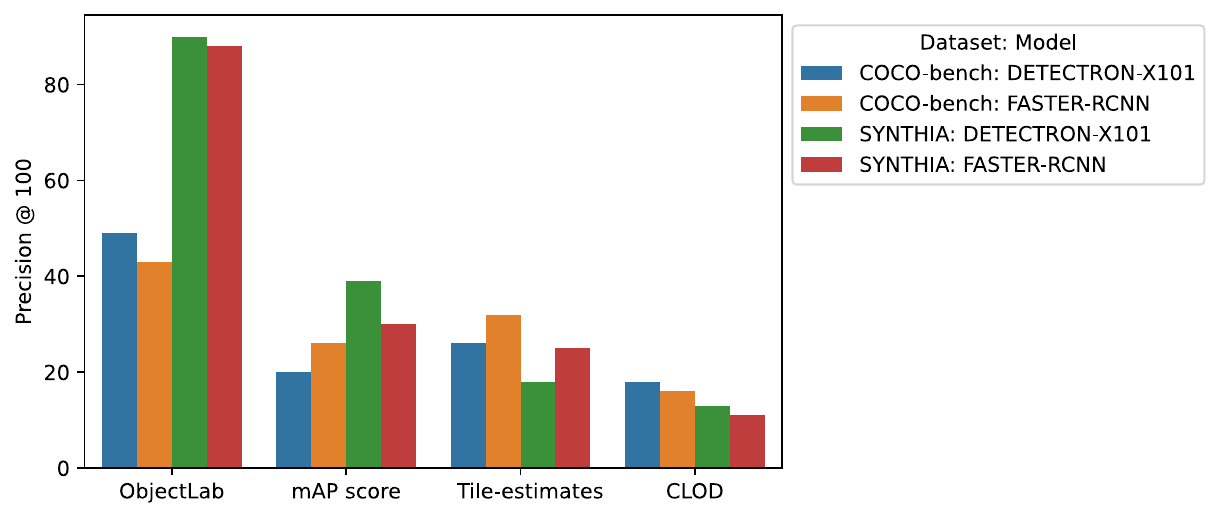} \hspace*{3mm} \\
 \includegraphics[width=85mm]{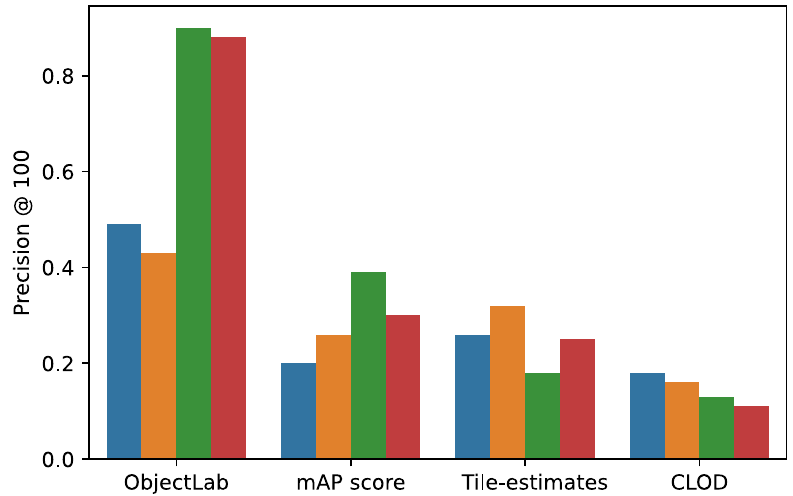} &   \includegraphics[width=85mm]{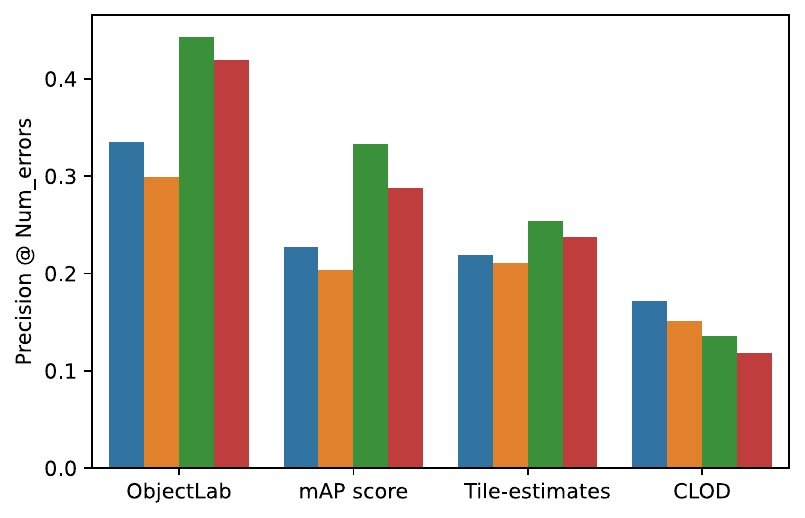} \\
\end{tabular}
\caption{Evaluating various label quality scoring methods across two models and two datasets where ground truth label errors are known.}
\label{fig:benchmark}
\end{figure*}

Figure \ref{fig:benchmark} shows that ObjectLab detects mislabeled images with  better precision/recall than other label quality scores in both COCO-bench and SYNTHIA-AL, regardless of which object detection model is used. Out of the other label quality scores evaluated, the basic mAP score performs the best on SYNTHIA-AL but does not fare as well on COCO-bench.


Table \ref{tab:fullcoco} evaluates our label quality score in the full 80-class COCO dataset. Because we could only calculate the precision via laborious manual review of top-ranking images under each method, we limit this evaluation to the \emph{Precision @ 100} metric and compare ObjectLab against the straightforward mAP label quality score. In COCO-full, ObjectLab again consistently detects label errors with much higher precision than the mAP score across both types of models.

\begin{table}[tb]
\caption{\emph{Precision@100} achieved by various label quality scores for detecting mislabeled images in the COCO-full dataset.}
\label{tab:fullcoco}
\vskip 0.15in
\begin{center}
\begin{small}
\begin{sc}
\begin{tabular}{lcccr}
\toprule
 Model & ObjectLab & mAP \\
\midrule
DETECTRON-X101  & 0.71 & 0.22 \\
FASTER-RCNN &  0.57 & 0.17\\
\bottomrule
\end{tabular}
\end{sc}
\end{small}
\end{center}
\vskip -0.1in
\end{table}

\begin{figure}[tb]
\vskip 0.2in
\begin{center}
\centerline{\includegraphics[width=\columnwidth]{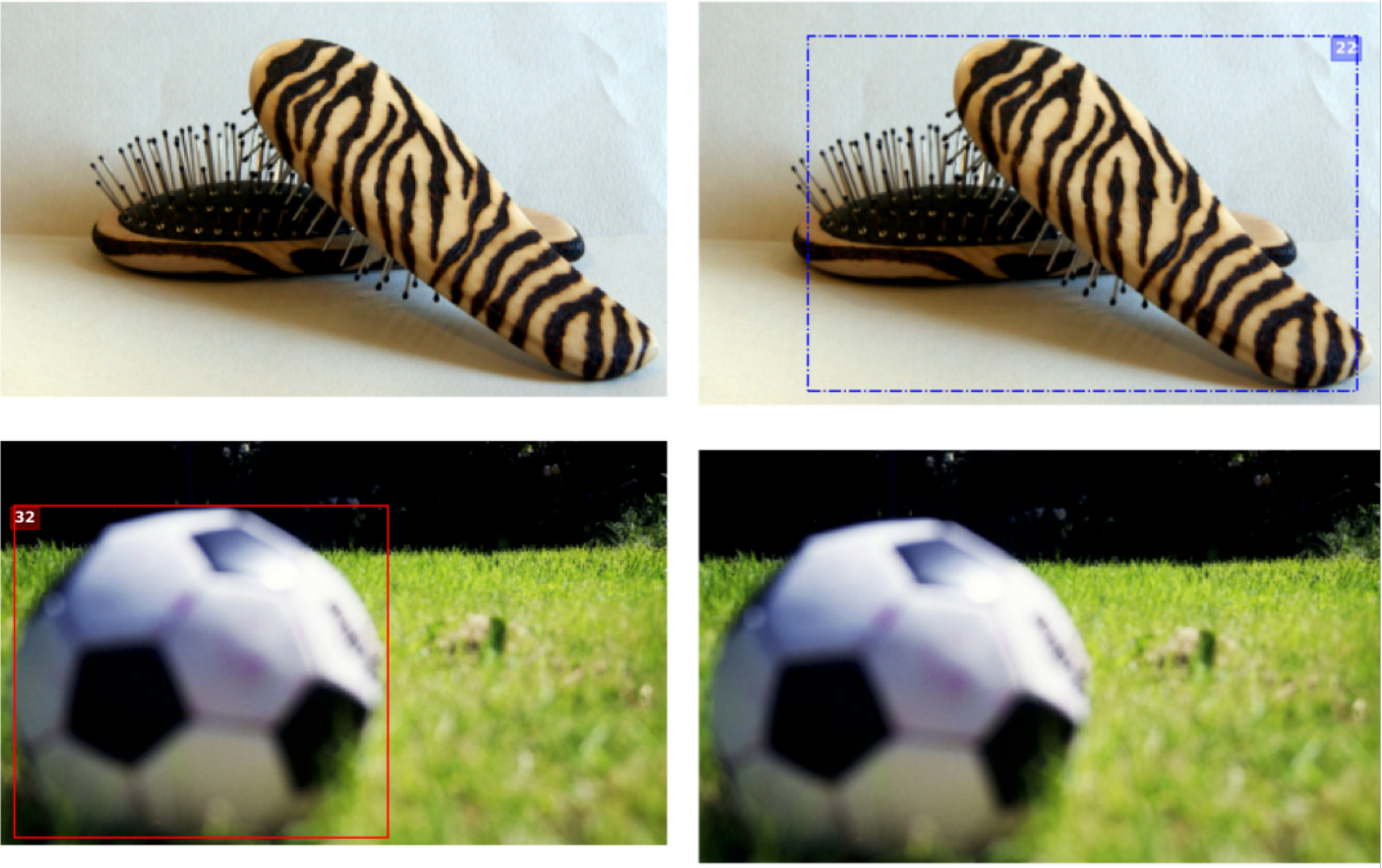}}
\caption{Images from COCO-full whose label quality is better assessed via ObjectLab than mAP score. We show both the original given label (left column in \textcolor{red}{red}) and prediction from Detectron2-X101 model (right column in \textcolor{blue}{blue}). 
\textbf{In top row}: model incorrectly predicts hairbrush is a zebra (class \#22) with moderate confidence. The label quality score for this image via mAP = 0.0, via ObjectLab = 1.0. 
\textbf{In bottom row}: model does not recognize sports ball (class \#32) but it is correctly marked in the annotation. The label quality score for this image via mAP = 0.0, via ObjectLab = 1.0. 
}
\label{fig:mapvs}
\end{center}
\vskip -0.2in
\end{figure}

While ObjectLab explicitly accounts for model confidence and specific forms of errors expected in practice, the mAP score exclusively relies on the IoU between predictions and labels.
Figure \ref{fig:mapvs} shows examples where mAP label quality scores mistakenly flag correctly labeled images due to imperfect predictions from the object detection model. Unlike mAP, ObjectLab scores are unaffected by bad model predictions made with insufficient confidence. Unlike mAP, ObjectLab scores do not explicitly penalize images with annotations for which there is no corresponding prediction. We did not encounter images with extraneously added bounding boxes in our examination of object detection datasets. 

\begin{figure*}[tb]
\centerline{\includegraphics[width=\textwidth]{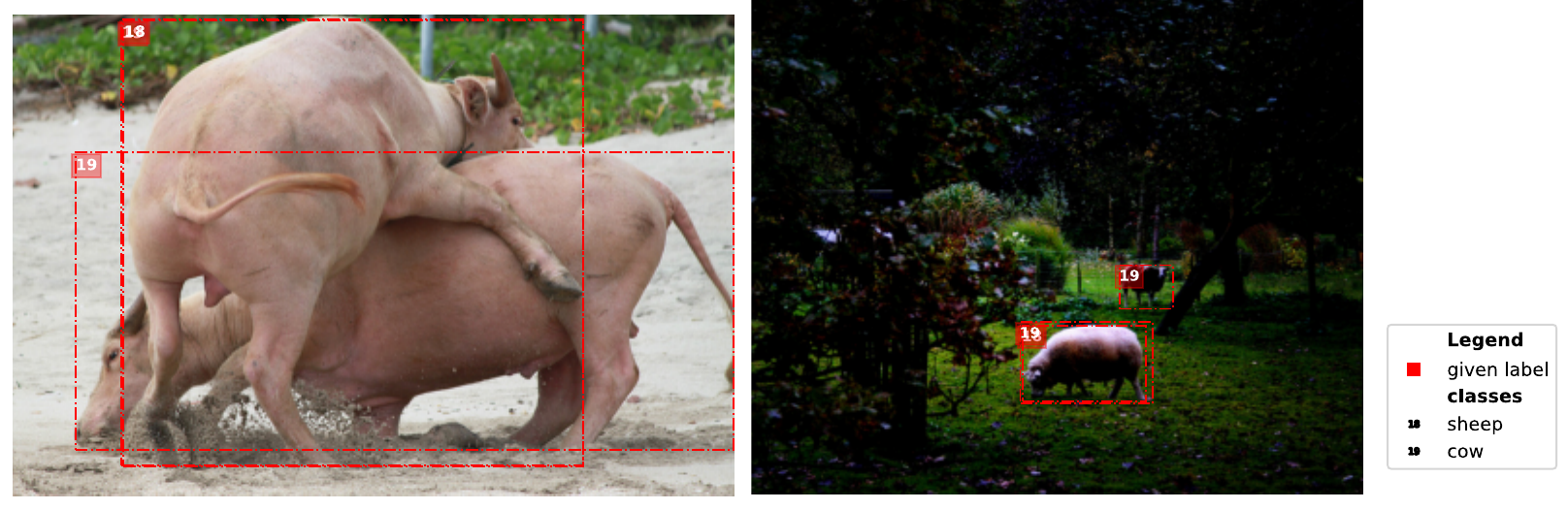}}
\caption{Examples of inconsistent annotations in COCO-full. In these images, which received low ObjectLab scores, some depicted animals are annotated as both \textbf{cow} and \textbf{sheep}.}
\label{fig:inconsistent}
\end{figure*}

The images with the lowest ObjectLab quality scores in the COCO-full dataset reveal many interesting findings. Figure \ref{fig:errortypes} illustrates different types of label errors present in the dataset, and Figure \ref{fig:inconsistent} shows fundamental inconsistencies between annotations. 
Through visual examination of many top/mid/bottom-ranking ObjectLab results, we estimate that in COCO 2017 around: 5\% of images have an \emph{Overlooked} error, 3\% have a \emph{Badly Located} error, and 0.7\% have a \emph{Swapped} error.

\section{Discussion}

The ObjectLab score introduced in this paper is straightforward to integrate into any existing object detection pipeline. Using predictions from the trained model, ObjectLab is able to detect diverse types of errors and automatically prioritizes mislabeled images in the data for review. After their labels are fixed, the same object detection model can be easily retrained on the corrected dataset. Because ObjectLab depends on model predictions, its label error detection accuracy increases with a better model. Thus a better model improves ObjectLab results, which in turn can be used to better correct the data, allowing an even better version of the model to be trained.  Most object detection models and datasets should be amenable to this virtuous cycle. In practice, labeling issues should be considered in both training and evaluation datasets \cite{northcutt2021pervasive}, to not only maximize reliability of a learned object detector but also ensure decisions like architecture selection and whether to deploy or not are based on the correct information.

\clearpage
\balance{}
\bibliography{objectlab}
\bibliographystyle{icml2023}
\balance{}

\newpage
\appendix
\onecolumn
\thispagestyle{empty}
\begin{center}
    {\LARGE \textbf{Supplementary Results}}
\end{center}


\vspace*{4em}

\begin{table}[htbp]
  \centering
  \begin{tabular}{lll}
    \toprule
    Dataset & Model & mAP \\
    \midrule
    COCO-bench  & X-101 & 0.67 \\
    COCO-bench  & FasterRCNN & 0.62 \\
    SYNTHIA  &  Detectron:X-101 & 0.58 \\
    SYNTHIA  & FasterRCNN & 0.53 \\
    COCO-full  & X-101 & 0.53 \\
    COCO-full  & FasterRCNN & 0.49 \\
    \bottomrule
  \end{tabular}
  \caption{Standard mAP evaluation to measure overall accuracy of the (out-of-sample)  predictions from each model on each dataset.}
\label{tab:accuracies}
\end{table}

\vspace*{4em}

\begin{table}[htbp]
  \centering
  \captionsetup{justification=centering, width=0.15\textwidth, font=small}
  \label{tab:metrics}

  \begin{tabular}{p{2.9cm}cccc}
    \toprule Dataset : Model & Quality Score & Avg. Precision & Precision@100 & Precision@Num\_errors \\
    \midrule
    \multirow{4}{*} & ObjectLab & 0.365 & 0.49 & 0.34 \\
    COCO-bench: & mAP & 0.222 & 0.20 & 0.23 \\
    X-101 & Tile-estimates & 0.284 & 0.26 & 0.22 \\
    & CLOD & 0.27 & 0.18 & 0.17 \\
    \midrule
    \multirow{4}{*} & ObjectLab & 0.273 & 0.43 & 0.30 \\
    COCO-bench:& mAP & 0.171 & 0.26 & 0.20 \\
    FRCNN & Tile-estimates & 0.283 & 0.32 & 0.21 \\
    & CLOD & 0.26 & 0.16 & 0.15 \\
    \midrule
    \multirow{4}{*} & ObjectLab & 0.502 & 0.89 & 0.44 \\
    SYNTHIA:& mAP & 0.313 & 0.39 & 0.33 \\
    X-101 & Tile-estimates & 0.280 & 0.18 & 0.25 \\
    & CLOD & 0.25 & 0.13 & 0.14 \\
    \midrule
    \multirow{4}{*} & ObjectLab & 0.46 & 0.88 & 0.42 \\
    SYNTHIA:& mAP & 0.309 & 0.30 & 0.29 \\
    FRCNN & Tile-estimates & 0.309 & 0.25 & 0.24 \\
    & CLOD & 0.23 & 0.11 & 0.12 \\
    \bottomrule
  \end{tabular}
  \caption{Metrics of various label quality scoring methods across two models and two datasets where ground truth label errors are known. This table is simply an alternate representation of the benchmark results plotted in Figure \ref{fig:benchmark}.}
\end{table}


\begin{figure}[ht]
\vskip 0.2in
\begin{center}
\centerline{\includegraphics[width=\columnwidth]{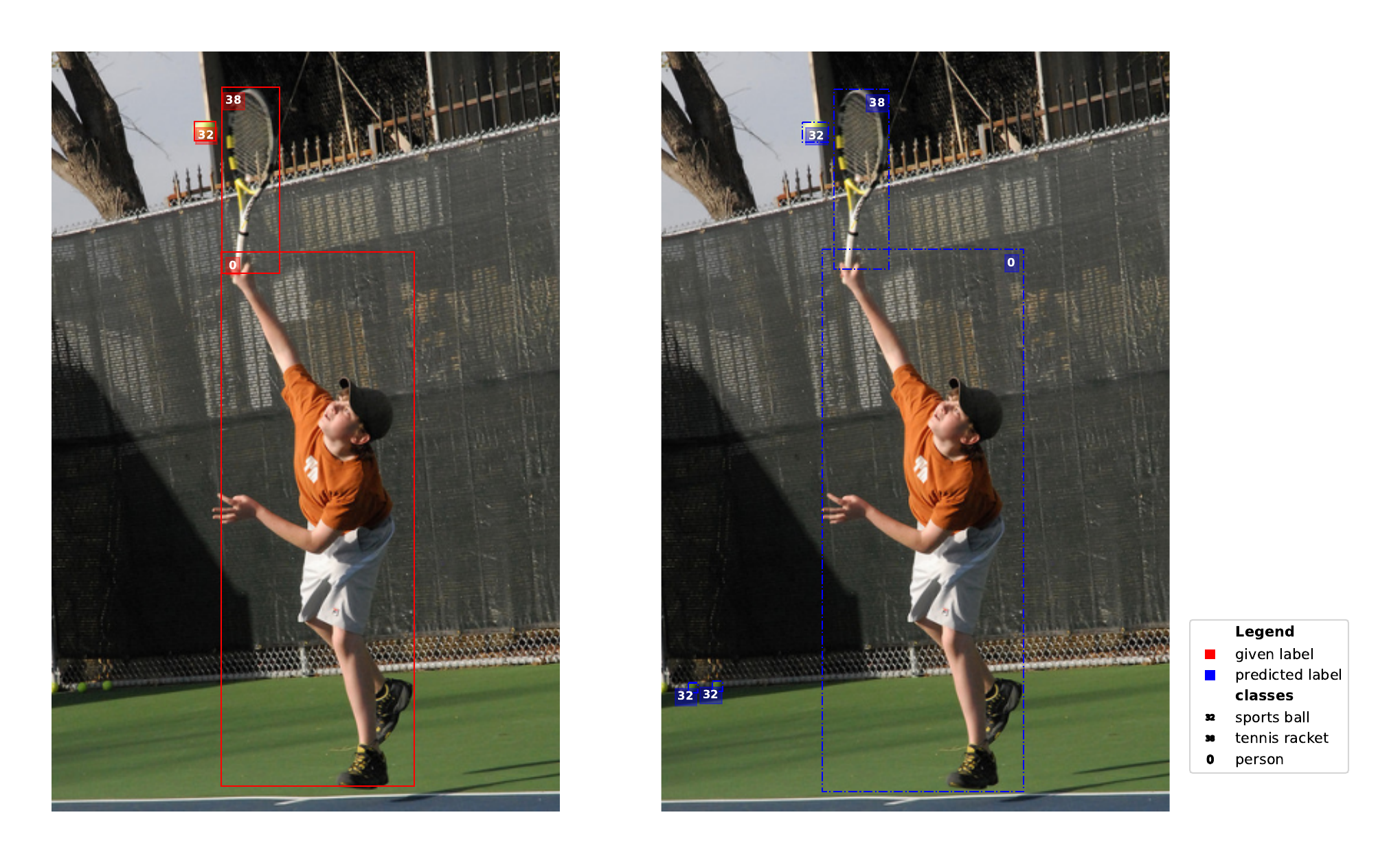}}
\centerline{\includegraphics[width=\columnwidth]{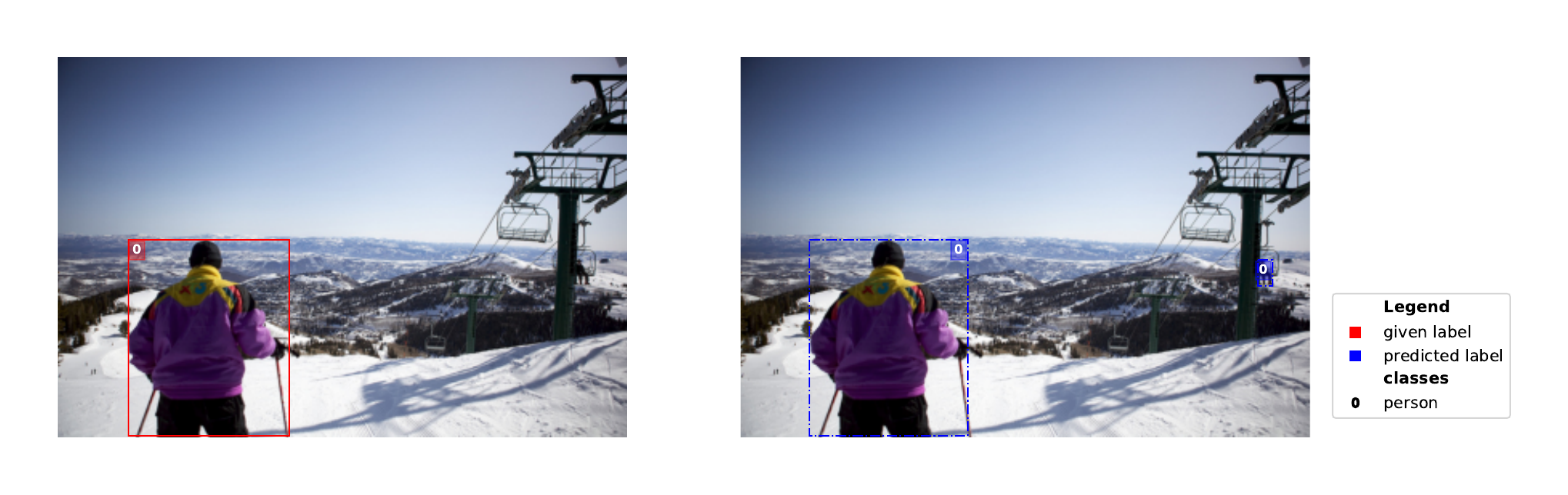}}
\caption{Additional examples of \emph{Overlooked} errors amongst the images with lowest ObjectLab scores in the COCO-full dataset. In top row: the given label (on left) is missing the \textbf{sport ball} (class \#32), detected by Detectron2-X101 model with confidence $ = 0.97$ (prediction on right). 
In bottom row: a \textbf{person} (class \#0) sitting on the chairlift was missed by annotators, but predicted by Detectron2-X101 model with confidence $ = 0.99$.}
\label{fig:moreoverlooked}
\end{center}
\vskip -0.2in
\end{figure}

\begin{figure}[ht]
\vskip 0.2in
\begin{center}
\centerline{\includegraphics[width=\columnwidth]{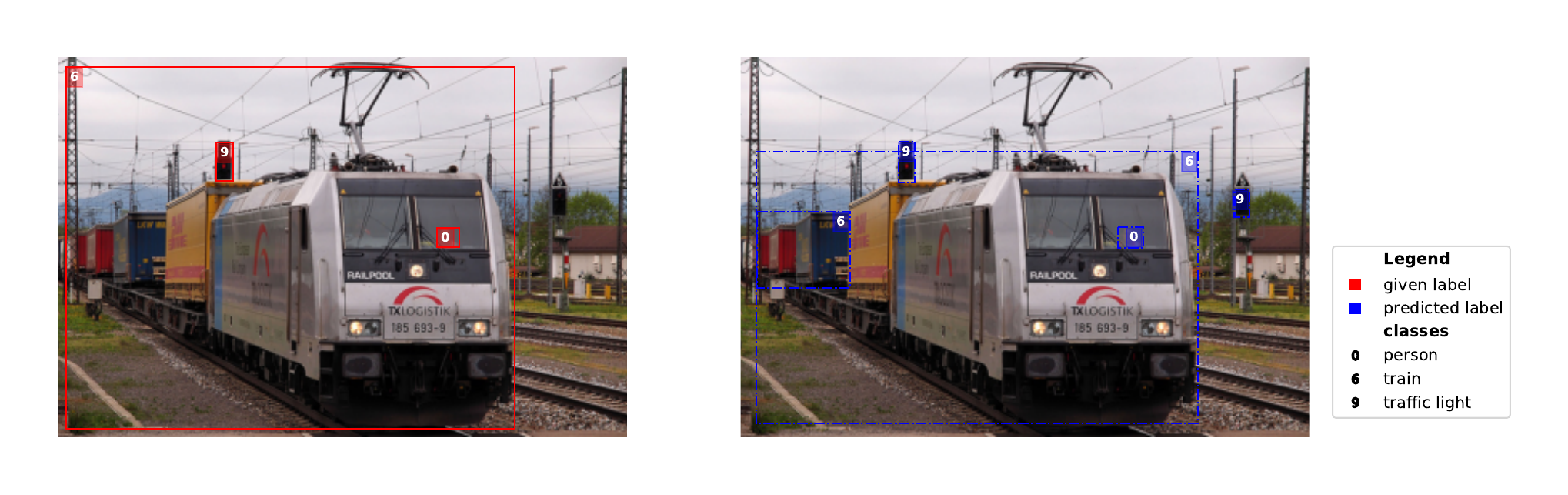}}
\centerline{\includegraphics[width=\columnwidth]{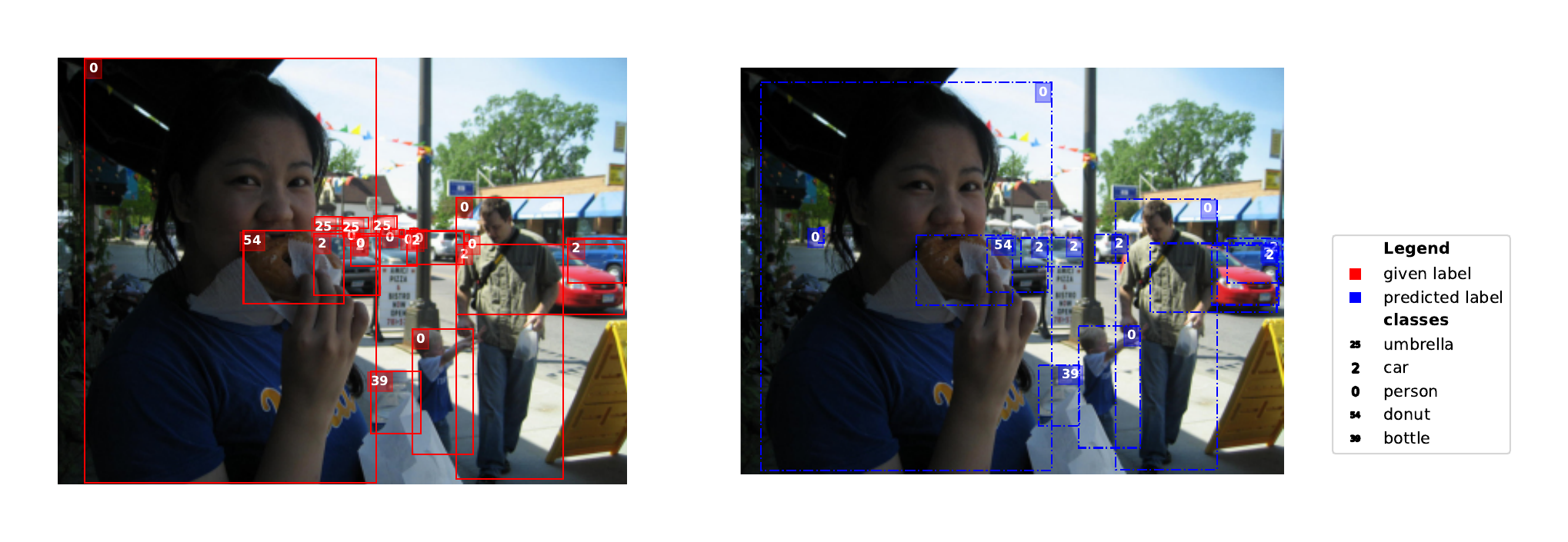}}
\caption{Additional examples of \emph{Badly Located} errors amongst the images with lowest ObjectLab scores in the COCO-full dataset. 
In top row: the given label (on left) incorrectly localizes the shape of the \textbf{train}, whereas Detectron2-X101 model correctly predicts its location with confidence $= 0.99$ (on right).
In bottom row: the annotated bounding box was poorly drawn around the \textbf{person} with extra room above their head.}
\label{fig:morebadloc}
\end{center}
\vskip -0.2in
\end{figure}


\begin{figure}[ht]
\vskip 0.2in
\begin{center}
\centerline{\includegraphics[width=\columnwidth]{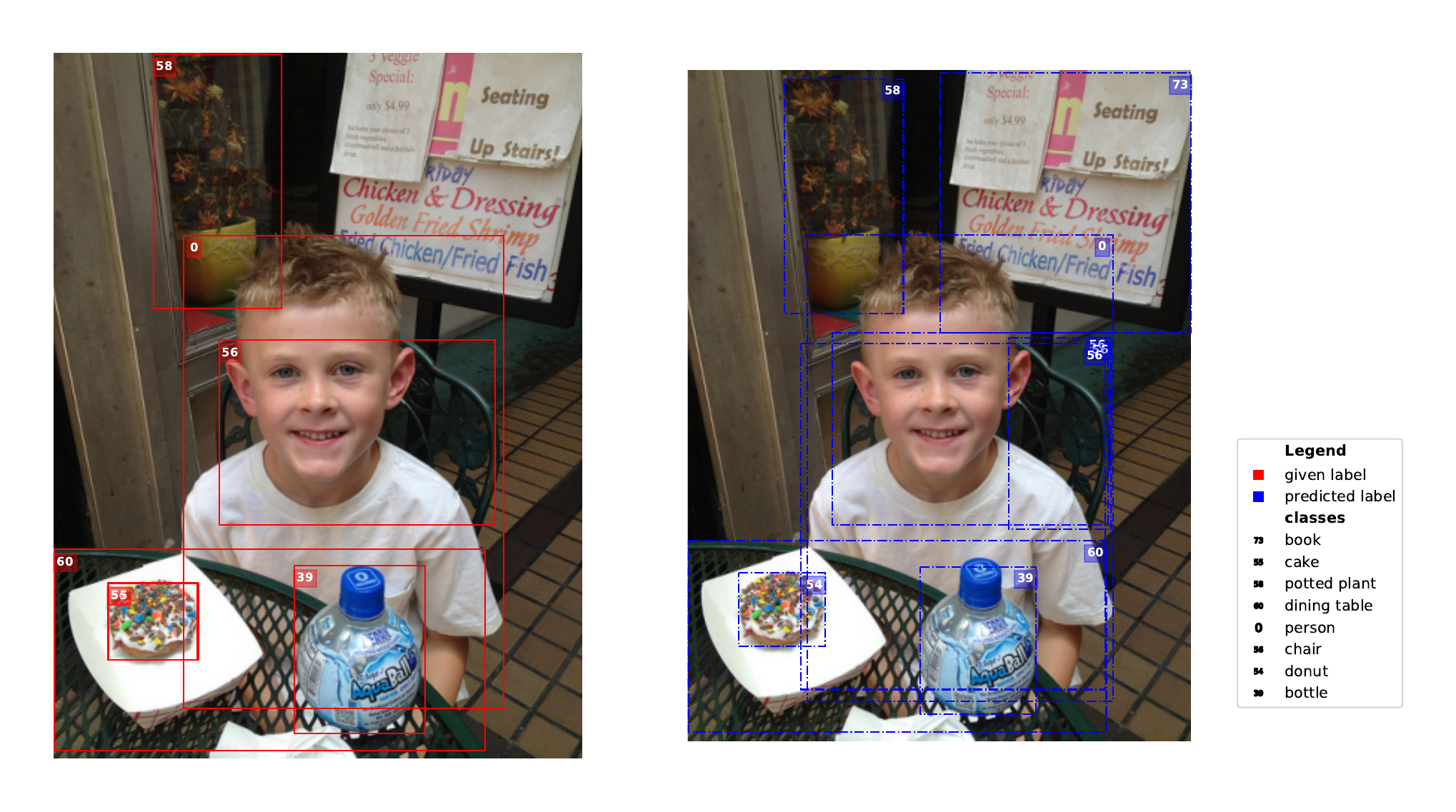}}
\centerline{\includegraphics[width=\columnwidth]{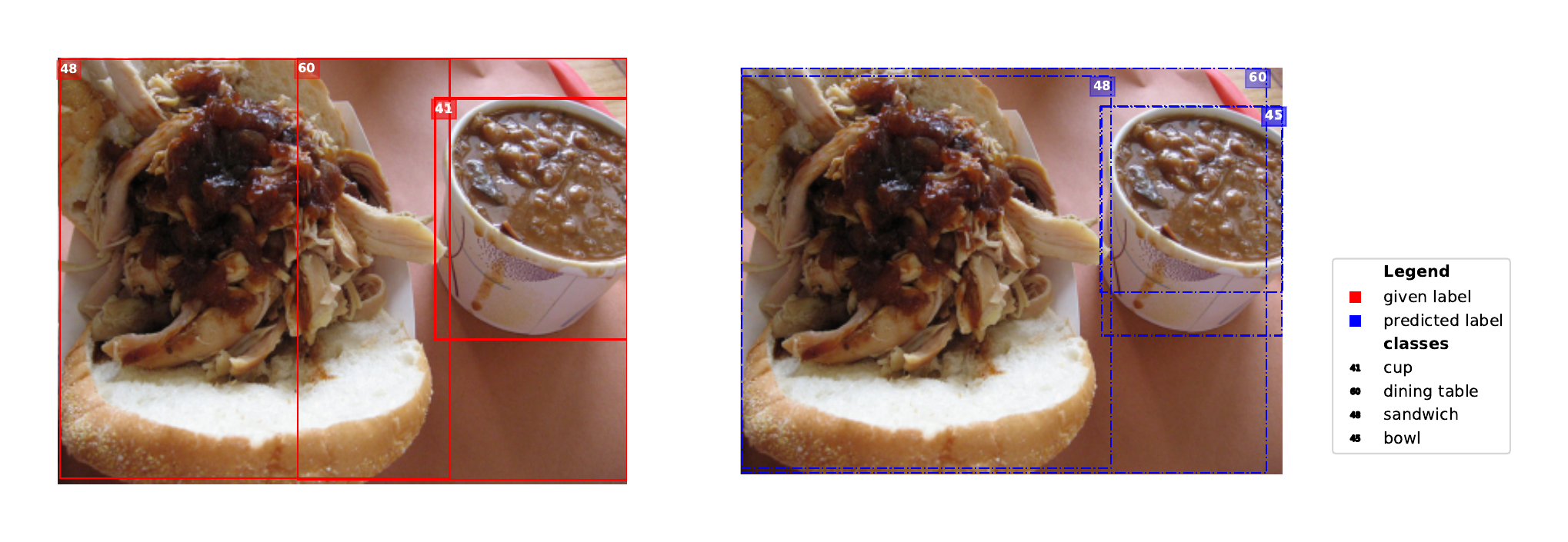}}
\caption{Additional examples of \emph{Swapped} errors amongst the images with lowest ObjectLab scores in the COCO-full dataset. 
In top row: the given label (on left) mistakenly says the dessert in the paper plate before the child is a \textbf{cake}, whereas Detectron2-X101 model predicts it is a \textbf{doughnut}.
In bottom row: the depicted \textbf{bowl} of chilly is incorrectly annotated as a \textbf{cup} in the given label.}
\label{fig:moreswapped}
\end{center}
\vskip -0.2in
\end{figure}


\begin{figure}[ht]
\begin{center}
\centerline{\includegraphics[width=\columnwidth]{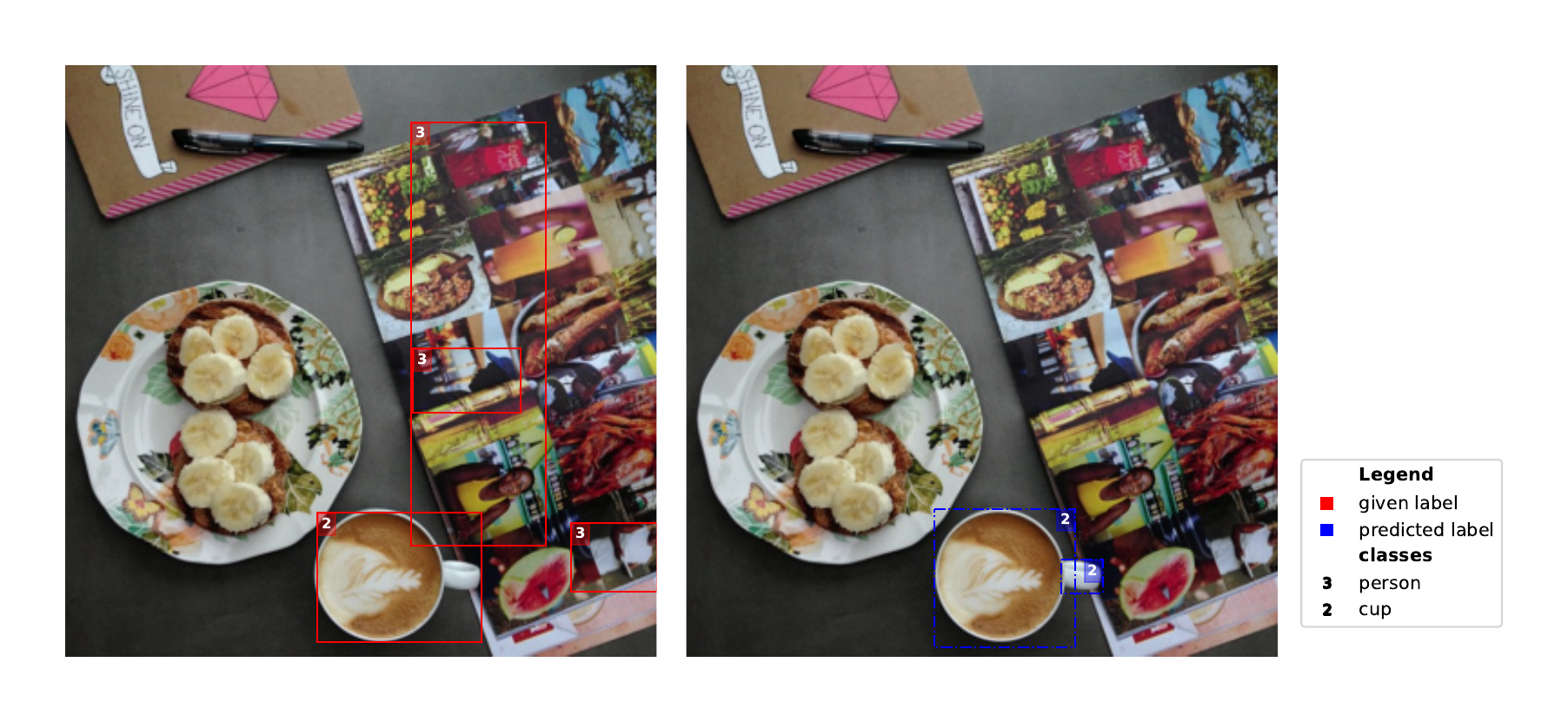}}

\centerline{\includegraphics[width=\columnwidth]{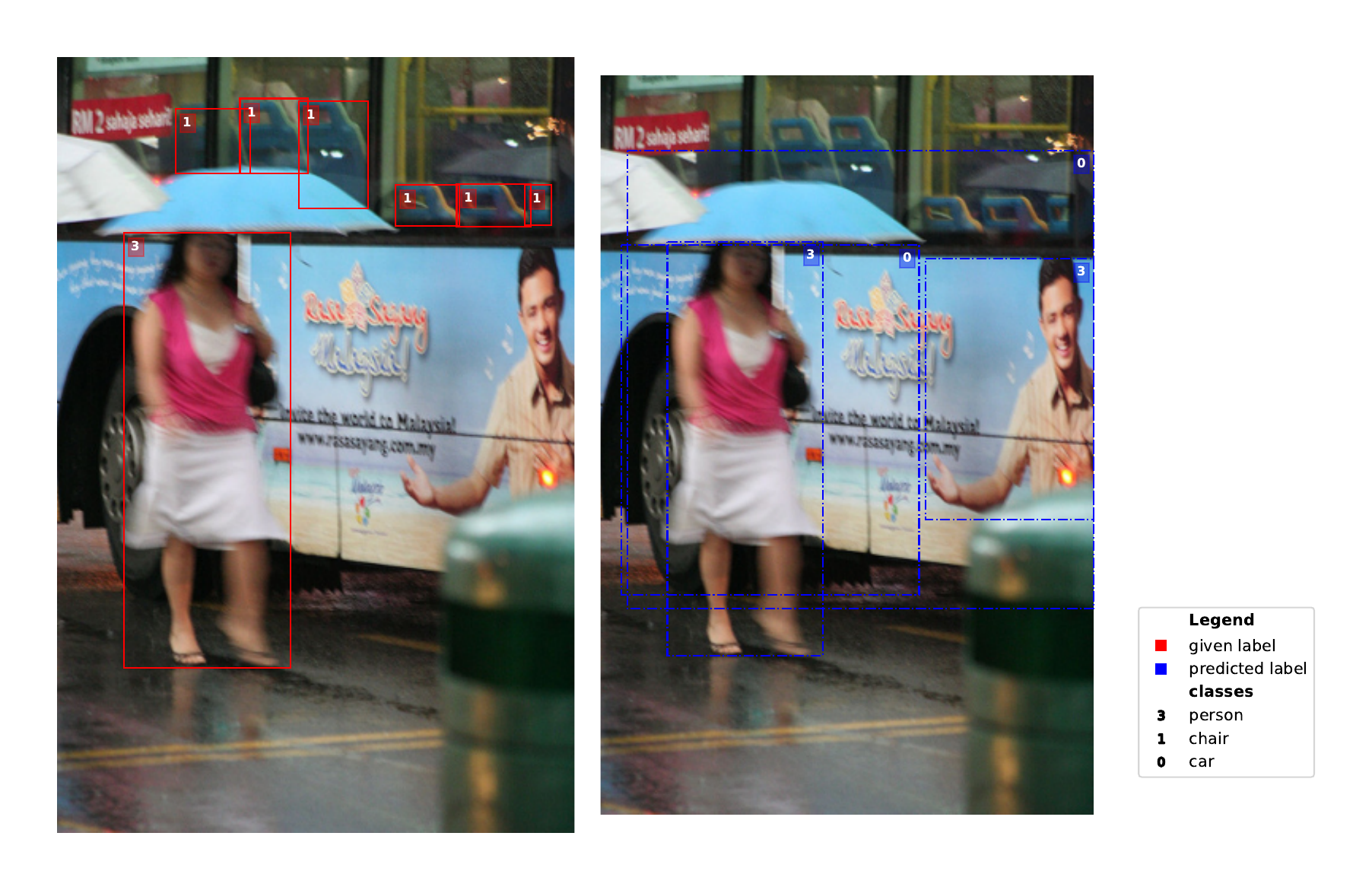}}

\caption{Examples of inconsistent annotations in COCO-full  detected via low ObjectLab score (with predictions from Detectron-X101 model). Here we see selective annotation of \emph{photos} of people as \textbf{person} objects in some images but not others.}
\label{fig:inconsistent2}
\end{center}
\end{figure}

\begin{figure}[ht]
\begin{center}
\centerline{\includegraphics[width=\columnwidth]{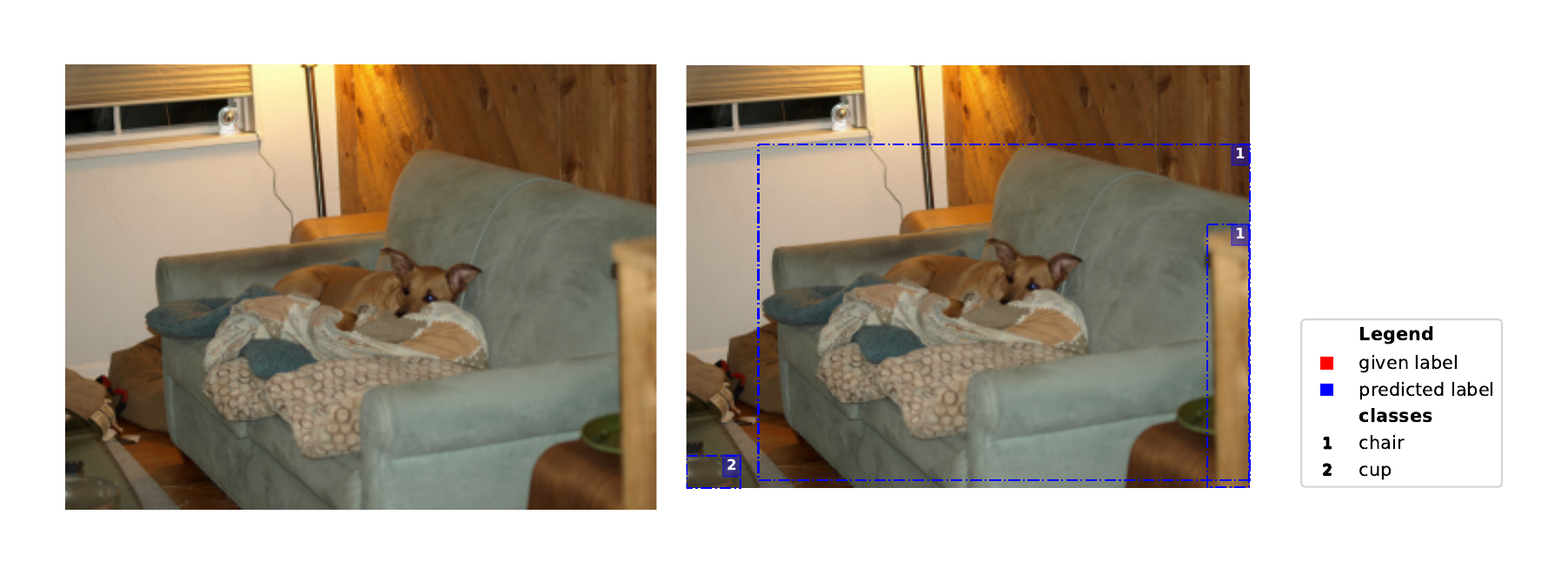}}
\centerline{\includegraphics[width=\columnwidth]{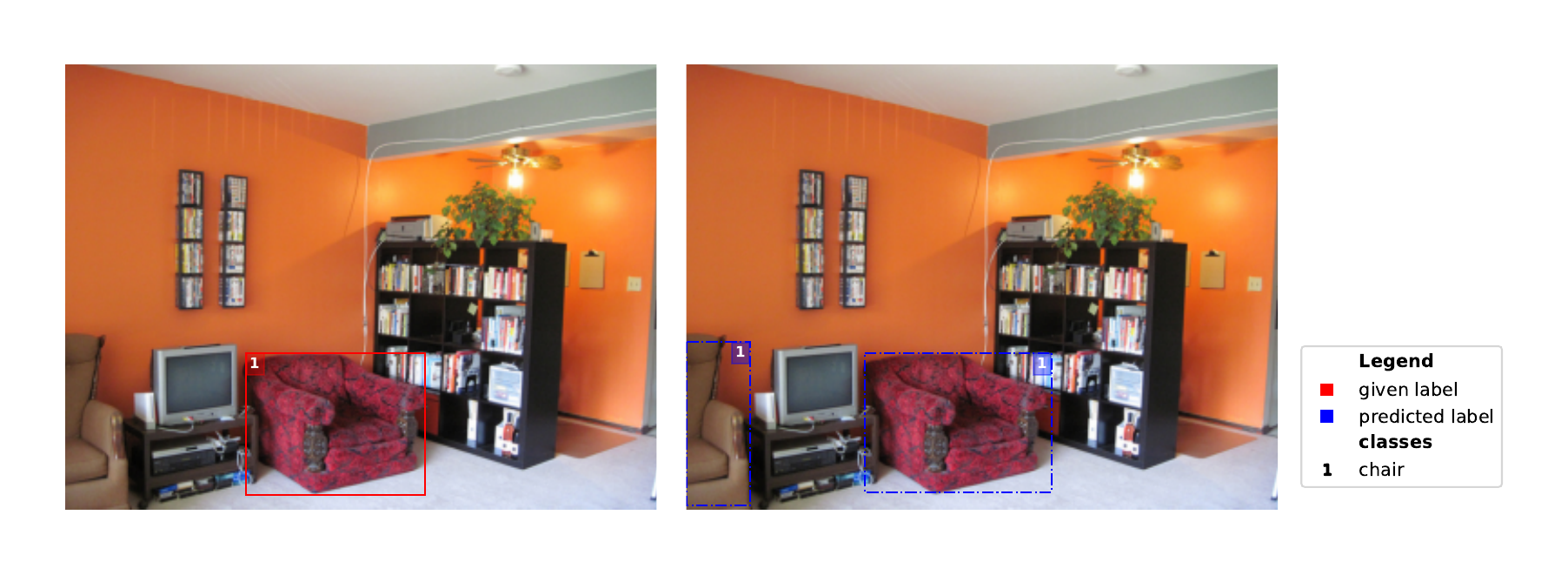}}
\caption{Examples of inconsistent annotations in COCO-full  detected via low ObjectLab score (with predictions from Detectron-X101 model). Here we see selective annotation of sofas as \textbf{chair} objects in some images but not others.}
\label{fig:inconsistent3}
\end{center}
\end{figure}

\begin{figure}[ht]
\begin{center}
\centerline{\includegraphics[width=\columnwidth]{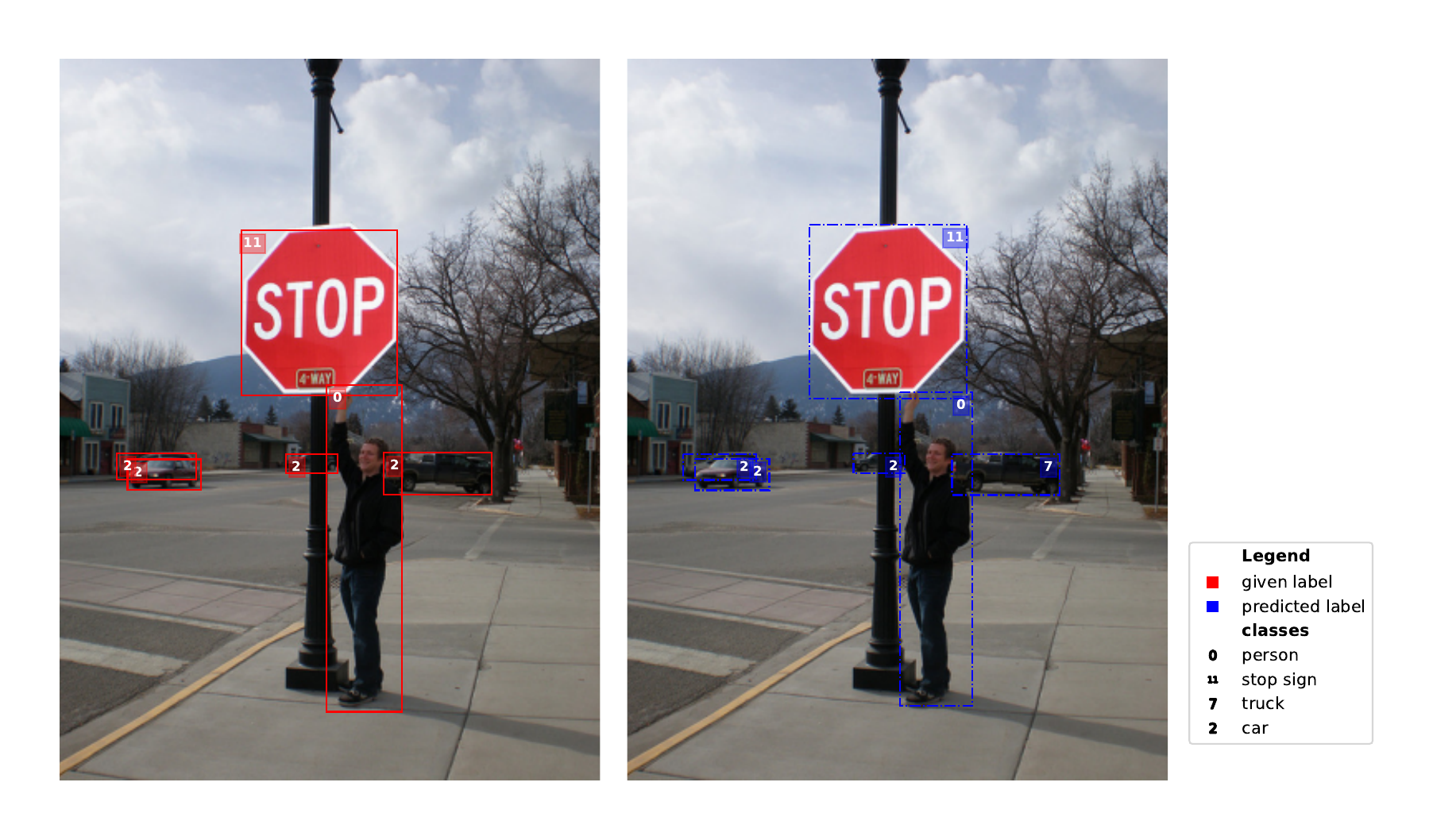}}
\centerline{\includegraphics[width=\columnwidth]{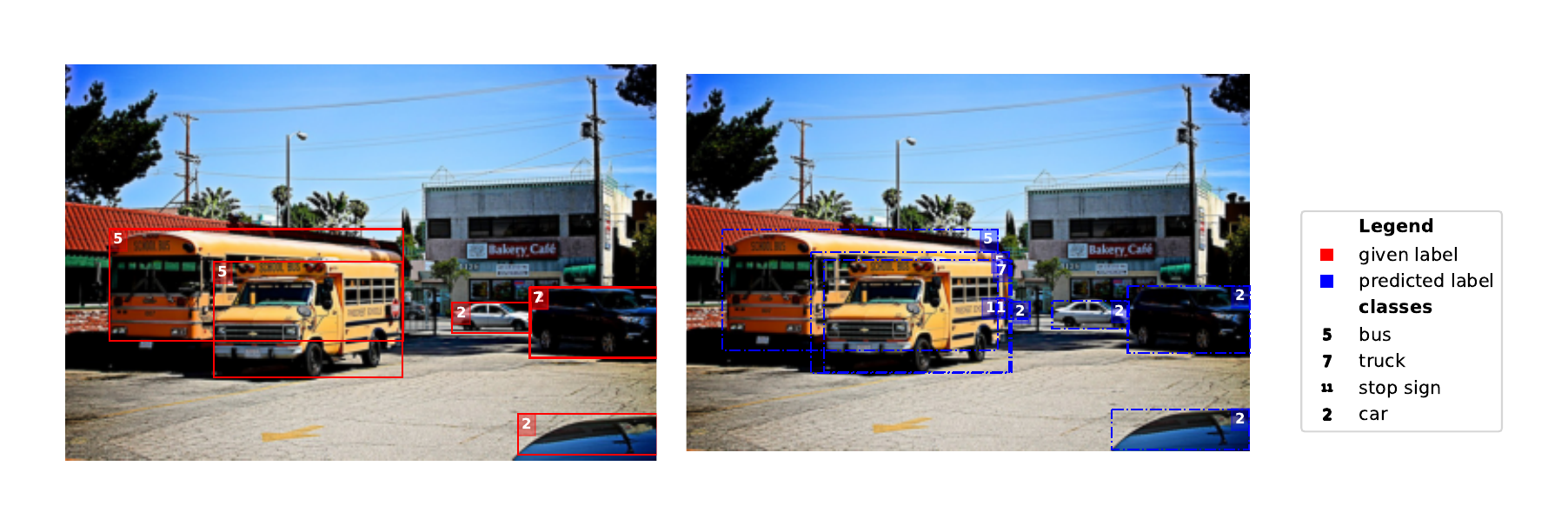}}
\caption{Examples of inconsistent annotations in COCO-full detected via low ObjectLab score (with predictions from Detectron-X101 model). Here we see selective annotation of SUVs as \textbf{car} objects in some images but as \textbf{truck} objects in others.}
\label{fig:inconsistent66}
\end{center}
\end{figure}

\begin{figure}[ht]
\begin{center}
\includegraphics[width=\columnwidth]{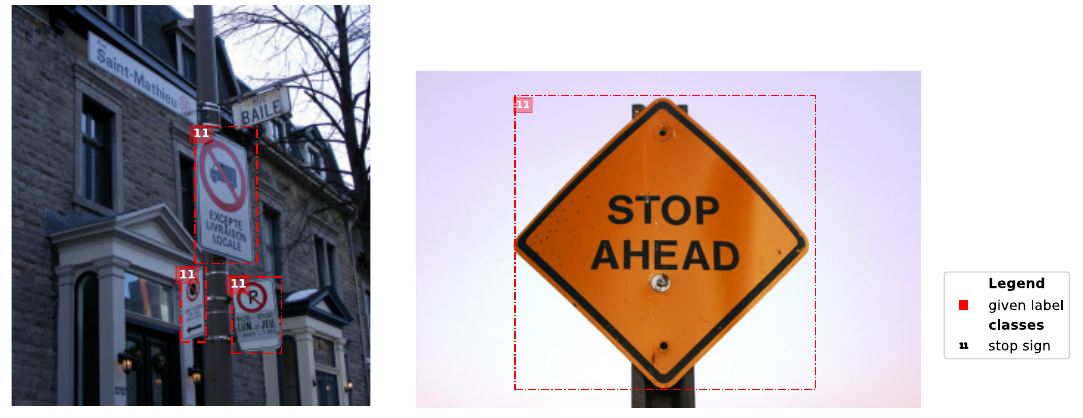}
\caption{More examples of common types of annotation mistakes in COCO-full that we discovered with the help of ObjectLab. Here various street signs are incorrectly annotated as \textbf{stop sign}.}
\label{fig:inconsistent1000}
\end{center}
\end{figure}

\begin{figure}[ht]
\begin{center}
\includegraphics[width=0.57\columnwidth]{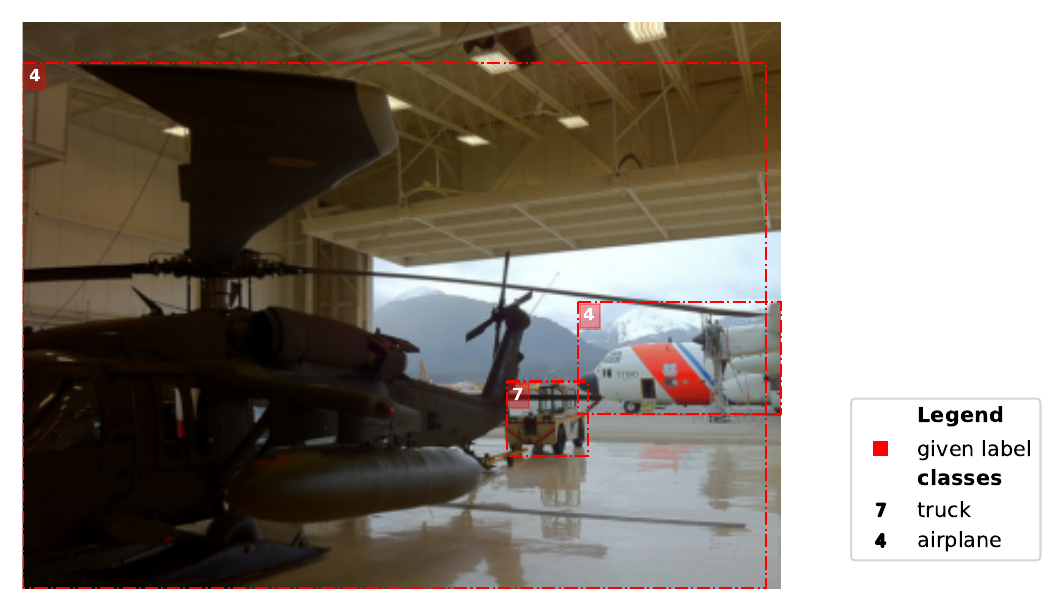}
\includegraphics[width=0.42\columnwidth]{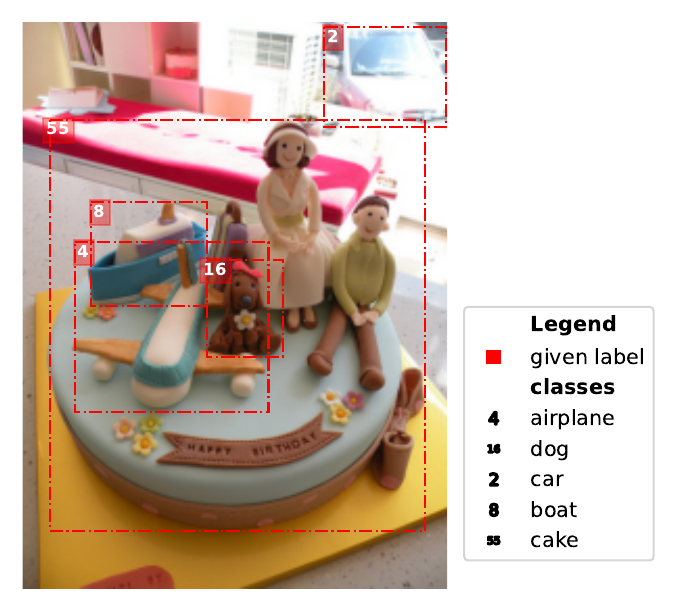}
\caption{More examples of common types of annotation mistakes in COCO-full that we discovered with the help of ObjectLab. On left: a helicopter is labeled as \textbf{airplane}, an error that sporadically occurs  throughout the dataset.
On right: certain toy figurines are annotated as the  \textbf{airplane}, \textbf{boat}, and \textbf{dog} objects they are imitations of, but the human figurines lack the \textbf{person} annotation. Such abstract representations of objects are sometimes annotated and sometimes not throughout the COCO dataset.
}
\label{fig:inconsistent_heli}
\end{center}
\end{figure}


\end{document}